\documentclass{article}
\usepackage{arxiv}

\usepackage{amsmath,amsfonts,bm}

\def\eqref#1{equation~\ref{#1}}

\def\1{\bm{1}}

\DeclareMathAlphabet{\mathsfit}{\encodingdefault}{\sfdefault}{m}{sl}
\SetMathAlphabet{\mathsfit}{bold}{\encodingdefault}{\sfdefault}{bx}{n}

\pdfoutput=1

\usepackage{microtype}
\usepackage{graphicx}
\usepackage{subcaption}
\usepackage{booktabs} 

\usepackage{makecell}
\usepackage{multirow}
\usepackage{xspace}
\usepackage{bm}
\usepackage{bbm}
\usepackage{tcolorbox}
\usepackage[normalem]{ulem}

\newcommand{\zzl}[1]{\textcolor{red}{#1}}

\usepackage{xcolor}
\definecolor{darkblue}{rgb}{0, 0.12, 0.55}
\definecolor{darkgreen}{rgb}{0, 0.55, 0.12}
\definecolor{darkred}{rgb}{0.6,0,0}
\definecolor{darkgreen}{rgb}{0,0.6,0}
\definecolor{Gray}{gray}{0.9}
\usepackage[breaklinks=true,
            colorlinks,
            linkcolor = darkred,
            urlcolor  = darkblue, 
            citecolor = teal,
            bookmarks = false]{hyperref}
\usepackage[numbers]{natbib}
\usepackage{url}

\usepackage{graphicx}
\usepackage{subcaption}
\usepackage{wrapfig}
\usepackage{booktabs}
\usepackage{multirow}
\usepackage{xspace}
\usepackage{array}
\usepackage{bm}
\usepackage{bbm}
\usepackage[ruled,vlined]{algorithm2e}
\usepackage{amsthm,amsmath,amssymb}

\usepackage{listings}
\usepackage{wrapfig}
\usepackage{subcaption}
\usepackage{colortbl}
\usepackage{adjustbox}
\usepackage{makecell}
\usepackage{pifont}
\usepackage{cleveref}
\usepackage{etoc}

\usepackage{algorithm}
\usepackage{algorithmic}

\definecolor{mymauve}{rgb}{0.58,0,0.82}
\makeatletter
\DeclareRobustCommand\onedot{\futurelet\@let@token\@onedot}
\def\@onedot{\ifx\@let@token.\else.\null\fi\xspace}

\def\eg{\emph{e.g}\onedot} 
 
\def\cf{\emph{c.f}\onedot} 
\def\etc{\emph{etc}\onedot}

\makeatother

\newlength{\mysize}

\allowdisplaybreaks

\theoremstyle{definition}

\title{DreamDPO: Aligning Text-to-3D Generation with \\ Human Preferences via Direct Preference Optimization}

\author{
Zhenglin Zhou$^{1}$ \ \ \ \ \
Xiaobo Xia$^{2\dagger}$ \ \ \ \ \ 
Fan Ma$^{3}$ \ \ \ \ \
Hehe Fan$^{1}$ \ \ \ \ \
Yi Yang$^{1\dagger}$ \ \ \ \ \
Tat-Seng Chua$^{2}$ \\
$^1$Zhejiang University \quad\quad
$^2$National University of Singapore \quad\quad
$^3$Yale University \\
\tt\small \{zhenglinzhou, hehefan, yangyics\}@zju.edu.cn \quad \{xbx, dcscts\}@nus.edu.sg \quad flowerfan524@gmail.com
}

\begin{document}

\maketitle
\def\thefootnote{$\dagger$}\footnotetext{Corresponding author.}

\begin{abstract}
Text-to-3D generation automates 3D content creation from textual descriptions, which offers transformative potential across various fields. However, existing methods often struggle to align generated content with human preferences, limiting their applicability and flexibility. To address these limitations, in this paper, we propose DreamDPO, an optimization-based framework that integrates human preferences into the 3D generation process, through direct preference optimization. Practically, DreamDPO first constructs pairwise examples, then compare their alignment with human preferences using reward or large multimodal models, and lastly optimizes the 3D representation with a preference-driven loss function. By leveraging pairwise comparison to reflect preferences, DreamDPO reduces reliance on precise pointwise quality evaluations while enabling fine-grained controllability through preference-guided optimization. Experiments demonstrate that DreamDPO achieves competitive results, and provides higher-quality and more controllable 3D content compared to existing methods. The code and models will be open-sourced. 
\href{https://zhenglinzhou.github.io/DreamDPO-ProjectPage/}{DreamDPO Webpage}.
\end{abstract}
\section{Introduction}
\label{sec:intro}
3D content generation is pivotal in driving innovation across diverse fields, including product design, medical imaging, scientific visualization, and the rapidly growing domains of virtual and augmented reality~\citep{li2023generative}. Despite its extensive applications, it remains challenging to create high-quality 3D content, which requires substantial time and effort, even for professionals. In response, \textit{text-to-3D generation} has emerged as a solution by automating 3D generation from textual descriptions, which archives remarkable advancements in the field~\citep{poole2022dreamfusion,wang2023prolificdreamer,wang2022sjc,yu2023text,wang2024prolificdreamer,shi2023mvdream,katzir2023noise,chung2023luciddreamer,wu2024consistent3d}. Nevertheless, some researchers~\citep{xie2024carve3d,ye2025dreamreward} emphasize that 3D content generated by existing methods often fails to align with \textit{human preferences} fully, highlighting the need for continued refinement and innovation in these methods.

Previous work~\citep{ye2025dreamreward} has leveraged \textit{reward models} to integrate human preferences into the generation process, leading to enhanced 3D generation outcomes. The core idea is to regularize the generated 3D content to achieve a high \textit{pointwise score} from the reward model. Despite these improved results, several issues remain to be addressed. First, it heavily depends on the reward model's ability to accurately evaluate the \textit{pointwise quality} of generated content, which places significant demands on the reward model. Second, since the reward model can only provide \textit{quality-relevant} scores, it lacks the flexibility to enable controllability from other perspectives. The issues reduce the applicability and adaptability of the current method. This falls short of meeting diverse requirements or providing broader control in 3D generation, which is never our desideratum.

To relieve the issues of prior work and better align 3D generation with human preferences, we propose DreamDPO. Essentially, DreamDPO is an optimization-based method for text-to-3D generation. It achieves alignment through direct preference optimization, leveraging preferences derived from either \textit{reward models} or \textit{large multimodal models}. Specifically, DreamDPO operates by initializing a 3D representation~\citep{mildenhall2020nerf,kerbl3Dgaussians} and optimizing it through a three-step iterative process. First, \textit{pairwise examples} are constructed \textit{on-the-fly} by applying different Gaussian noise. Second, a reward model or a large multimodal model \textit{ranks} these examples based on their matching with the input text prompt or \textit{specific instructions}, which matches human preferences about the pairwise examples\footnote{Reward models are trained using pairwise data reflecting human preferences, while large multimodal models are inherently aligned with these preferences~\citep{sun2023aligning}.}. Finally, a reward loss is computed from the \textit{pairwise preferences}, which guides the update of the 3D representation. By incorporating human preferences into the optimization loop, DreamDPO generates 3D assets that achieve superior alignment with textual inputs, along with enhanced texture and geometry quality.

DreamDPO can be justified as follows. While absolute quality evaluation inherently provides a ranking on pairwise examples, a ranking requires only the scores to distinguish \textit{relative preferences}, but need not be perfectly accurate~(\textit{c.f.}, \citep{zhang2024generating}). DreamDPO takes advantage of this distinction, and changes the previous \textit{score-guided} optimization~\citep{ye2025dreamreward} to \textit{preference-guided} optimization. Therefore, it lowers the demand for precise scoring and requires only distinguishable scores. Additionally, DreamDPO can make use of the preferences provided by large multimodal models. By constructing preferred and less preferred examples based on specific instructions about the attributes of generation content~(\eg., the object number and motion), it directs the optimization process to align more closely with the desired outcomes.  This strategy enhances adherence to instructions and introduces \textit{fine-grained controllability}, meeting diverse requirements effectively. Moreover, we conduct a series of experiments to justify our claims. Empirical results demonstrate that our method flexibly accommodates either reward models or large multimodal models, which enables the generation of higher-quality and more controllable 3D content. 

Before delving into details, we clearly emphasize our contribution as follows.
\begin{itemize}
    \item Conceptually, we propose DreamDPO that shifts the paradigm of text-to-3D generation by reducing dependence on precise absolute quality evaluation. Instead, it leverages distinguishable relative preferences and integrates large multimodal models, which achieves remarkable alignment with human preferences while enabling fine-grained controllability.
    \item Technically, DreamDPO pioneers a three-step optimization process, combining online pair construction, preference ranking, and ranking-driven updates. This innovative design ensures superior alignment with input prompts, significantly enhancing the quality and adaptability of generated 3D content.
    \item Empirically, extensive results establish DreamDPO as a new benchmark for text-to-3D generation. Both quantitative and qualitative results are thoroughly analyzed, supported by detailed discussions and ablation studies. DreamDPO outperforms 13 state-of-the-art methods, achieving the best quantitative performance across two key metrics while delivering highly impressive and controllable qualitative results.
\end{itemize}

The rest of this paper is organized as follows. Section~\ref{sec:pre} introduces some background knowledge relevant to this work. Section~\ref{sec:method} presents the technical details of our proposed DreamDPO. Experimental results are analyzed and discussed in Section~\ref{sec:exp}. Conclusions are given in Section~\ref{sec:clu}.

\begin{figure*}[t]
    \centering
    \includegraphics[width=1.0\linewidth]{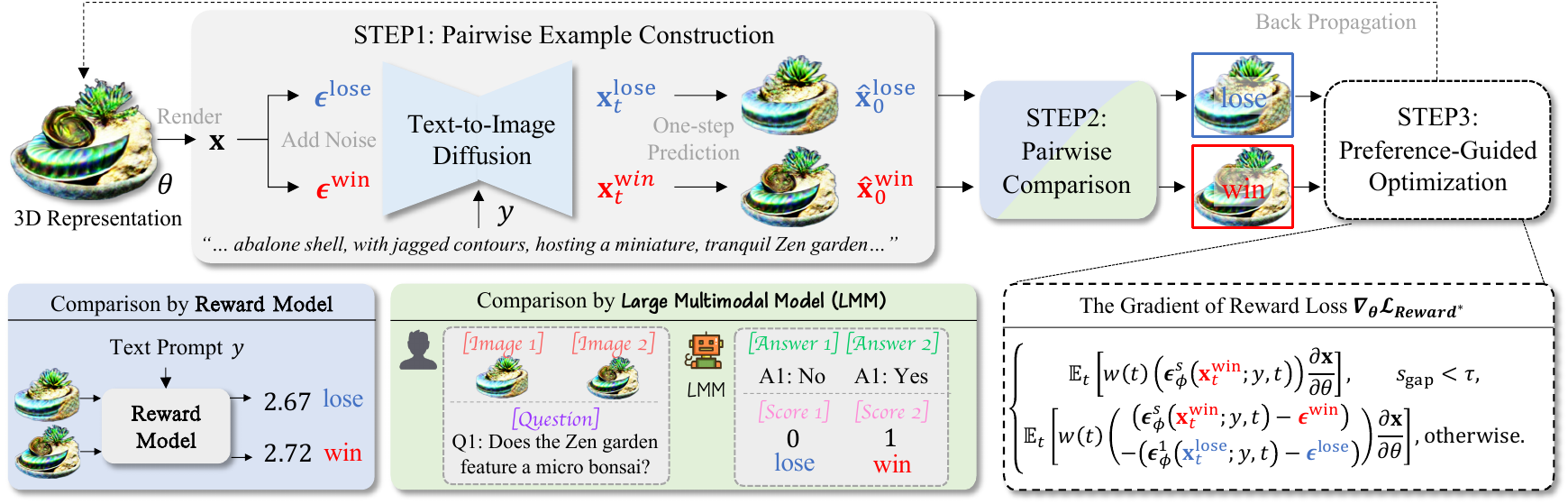}
    \caption{Overview of our method.
    DreamDPO first constructs pairwise examples, then compares their alignment with human preferences using reward or large multimodal models, and lastly optimizes the 3D presentation with a preference-driven loss function. The loss function pulls the \texttt{win} example $\mathbf{x}_t^{\text{win}}$ closer and pushes the \texttt{lose} example $\mathbf{x}_t^{\text{lose}}$ away. 
    As a piecewise objective, it selectively pushes $\mathbf{x}_t^{\text{lose}}$ only when the preference score gap $s_\text{gap}$ exceeds a threshold $\tau$, preventing chaotic gradients from overly similar $\mathbf{x}_t^{\text{lose}}$.
    }
    \label{fig:overview}
\end{figure*}
\vspace{-7pt}
\section{Preliminaries}
\label{sec:pre}
\vspace{-3pt}
Text-to-3D generation aims to create high-quality 3D assets aligned with a given text prompt $y$.
The pipeline typically distills knowledge from a parametrized diffusion model $\bm{\epsilon}_\phi$~\citep{rombach2022high,shi2023mvdream} into a learnable 3D representation with parameters $\theta\in\Theta$ (\eg, NeRF~\citep{mildenhall2020nerf}, DMTet~\citep{shen2021deep}, and 3DGS~\citep{kerbl3Dgaussians}), where $\Theta$ is the space of $\theta$ with the Euclidean metric. Score distillation sampling~(SDS)~\citep{poole2022dreamfusion} is used to guide the distillation process. 

\textbf{Diffusion models.} The diffusion model has been widely used in generative tasks~\citep{sohl2015deep,song2022diffusion,bao2022analytic,peebles2023scalable}. Generally, it involves a forward process to gradually add Gaussian noise to data points and a reverse process to transform Gaussian noise into data points from a target distribution $p_{\mathrm{data}}$. The reverse process starts from an initial noise $\mathbf{z}_T\sim\mathcal{N}(\mathbf{0}, \mathbf{I})$. At each diffusion step $t$, the model refines noisy data $\mathbf{z}_t$ into a cleaner one $\mathbf{z}_{t-1}$ until finally producing $\mathbf{z}_0 = \mathbf{x}\sim p_{\mathrm{data}}$. Therefore, the transitions $p(\mathbf{z}_{t-1}|\mathbf{z}_{t})$ can be learned effectively by the diffusion model.

\textbf{Score distillation sampling~(SDS).} SDS was proposed in DreamFusion~\citep{poole2022dreamfusion}, and has been widely studied~\citep{wang2023prolificdreamer,yu2023text,katzir2023noise,chung2023luciddreamer,wu2024consistent3d,zhuo2025vividdreamer,ye2025dreamreward}. Technically, for a rendered image $\mathbf{x}$ from a 3D representation, random noise $\bm{\epsilon}$ is added at timestep $t$. 
A pre-trained diffusion model predicts this noise. The SDS loss is computed as the difference between predicted and added noise, which optimizes a set of parameters $\theta$. The gradient of the SDS loss with respect to $\theta$ is:
\begin{equation}
    \nabla_\theta \mathcal{L}_{\text{SDS}} = \mathbb{E}_{t, \bm{\epsilon}}[w(t) (\bm{\epsilon}^s_{\phi}(\mathbf{x}_t;y,t)-\bm{\epsilon})\frac{\partial\mathbf{x}}{\partial\theta}],
\end{equation}
where $\mathbf{x}_t = \alpha_t \mathbf{x}_0 + \sigma_t \bm{\epsilon}$, $w(t)$ is a weighting function, and $s$ is a pre-defined scalar of classifier-free guidance (CFG)~\citep{ho2022classifier}. The minimization of the SDS loss follows the score function of the diffusion model to move $\mathbf{x}$ to the text description region, ensuring the generated 3D representation aligns with the given text prompt. Note that to make a continuous and smooth presentation, we provide a detailed review of related work in Appendix~\ref{app:related_work}.

\section{Method}
\label{sec:method}

\textbf{Overview.} DreamDPO is an optimization-based text-to-3D generation method. It begins by initializing a 3D representation, \eg, NeRF~\citep{mildenhall2020nerf}. In each training iteration, the optimization procedure involves three key steps: (1) \textit{pairwise example construction}: pairwise examples are generated online by applying different Gaussian noises during the diffusion process; (2) \textit{pairwise example comparison}: a reward model or a large multimodal model~(LMM) compares the generated examples based on their alignment with the desired text prompt; and (3) \textit{preference-guided optimization}: a piecewise reward loss is calculated using the pairwise comparison, and the 3D representation is updated accordingly. Overall, our DreamDPO guides the optimization process with human preferences, leading to 3D assets with improved alignment to input text and enhanced texture/geometry quality. The framework overview of our method is provided in \cref{fig:overview}. A complete algorithm flow of our method can be checked in Appendix~\ref{app:algorithm_flow}. 

\begin{table*}[t] 
    \centering
    \caption{
    Qualitative comparisons on 110 prompts generated by GPTEval3D~\protect\citep{wu2024gpt}.
    We calculate the ImageReward score (IR)~\protect\citep{xu2024imagereward} for human preference evaluation, the CLIP score~\protect\citep{radford2021clip} for text-image alignment evaluation, and GPTEval3D~\protect\citep{wu2024gpt} for comprehensive 3D quality evaluation. The best performance in each case is shown in bold.
    }
    \begin{tabular}{c|ccccccc} 
    \toprule
    \multirow{2}{*}{Method} & \multirow{2}{*}{IR $\uparrow$}  & \multicolumn{6}{c}{GPTEval3D $\uparrow$} \\ \cmidrule{3-8}
    & & \footnotesize Alignment & \footnotesize Plausibility & \scriptsize T-G Coherency. & \footnotesize Geo Details &  \footnotesize Tex Details & \footnotesize Overall \\
    \midrule
    DreamFusion~\citep{poole2022dreamfusion} & -1.51 & 1000.0 & 1000.0 & 1000.0 & 1000.0 & 1000.0 & 1000.0 \\
    DreamGaussian~\citep{tang2023dreamgaussian} & -1.56 & 1100.6 & 953.6 & 1158.6 & 1126.2 & 1130.8 & 951.4 \\
    Fantasia3D~\citep{chen2023fantasia3d} & -1.40 & 1067.9 & 891.9 & 1006.0 & 1109.3 & 1027.5 & 933.5 \\
    Instant3D~\citep{li2023instant3d} & -0.91 & 1200.0 & 1087.6 & 1152.7 & 1152.0 & 1181.3 & 1097.8 \\
    Latent-NeRF~\citep{metzer2023latent} & -0.42  & 1222.3 & 1144.8 & 1156.7 & 1180.5 & 1160.8 & 1178.7 \\
    Magic3D~\citep{lin2023magic3d} & -1.11 & 1152.3 & 1000.8 & 1084.4 & 1178.1 & 1084.6 & 961.7 \\
    Point-E~\citep{nichol2022point} & -2.24  & 725.2 & 689.8 & 688.6 & 715.7 & 745.5 & 618.9 \\
    ProlificDreamer~\citep{wang2023prolificdreamer} & -0.50 & 1261.8 & 1058.7 & 1152.0 & 1246.4 & 1180.6 & 1012.5 \\
    Shap-E~\citep{jun2023shap} & -2.10 & 842.8 & 842.4 & 846.0 & 784.4 & 862.9 & 843.8 \\
    SJC~\citep{wang2023score} & -0.82 & 1130.2 & 995.1 & 1033.5 & 1079.9 & 1042.5 & 993.8 \\
    SyncDreamer~\citep{liu2023syncdreamer} & -1.77 & 1041.2 & 968.8 & 1083.1 & 1064.2 & 1045.7 & 963.5 \\
    Wonder3D~\citep{long2024wonder3d} & -1.70 & 985.9 & 941.4 & 931.8 & 973.1 & 967.8 & 970.9 \\
    MVDream~\citep{shi2023mvdream} & -0.58 & 1270.5 & 1147.5 & 1250.6 & 1324.9 & 1255.5 & 1097.7 \\
    \midrule
    DreamDPO~(ours) & \textbf{-0.35} & \textbf{1298.9} & \textbf{1171.9} & \textbf{1276.4} & \textbf{1373.2} & \textbf{1296.9} & \textbf{1203.1} \\
\bottomrule
    \end{tabular}
    \label{tab:qua_comp}
\end{table*}

\subsection{Algorithm Details}

\textbf{Pairwise example construction.} Given a sampled camera pose, an RGB image $\mathbf{x}$ can be rendered from the 3D representation using renderers. Then two different Gaussian noise $\bm{\epsilon}^1$ and $\bm{\epsilon}^2$ are added to $\mathbf{x}$ at timestep $t$, resulting in pairwise noisy images $\mathbf{x}_t^1$ and $\mathbf{x}_t^2$:
\begin{equation}
    \mathbf{x}_t^1 = \alpha_t \mathbf{x}_0 + \sigma_t \bm{\epsilon}^1, \ \ \mathbf{x}_t^2 = \alpha_t \mathbf{x}_0 + \sigma_t \bm{\epsilon}^2,
\end{equation}
where $\mathbf{x}_0 = \mathbf{x}$, $\alpha_t$ and $\sigma_t$ are hyperparameters satisfying $\alpha_0 \approx 1, \sigma_0 \approx 0, \alpha_0 \approx 0, \sigma_0 \approx 1$~(\cf, \citep{sohl2015deep,ho2020denoising}). Afterward, we feed the pairwise noisy images into a pre-trained text-to-image diffusion model $\bm{\epsilon}_\phi$~\citep{shi2023mvdream,rombach2022high} and generate corresponding predictions:
\begin{equation}
    \begin{aligned}
       \hat{\mathbf{x}}_0^1 = \frac{\mathbf{x}_t^1 - \sqrt{1 - \alpha_t}\bm{\epsilon}_\phi(\mathbf{x}_t^1;y,t)}{\sqrt{\alpha_t}},  \\
       \hat{\mathbf{x}}_0^2 = \frac{\mathbf{x}_t^2 - \sqrt{1 - \alpha_t}\bm{\epsilon}_\phi(\mathbf{x}_t^2;y,t)}{\sqrt{\alpha_t}},
    \end{aligned}
\end{equation}
where $\hat{\mathbf{x}}_0^1$ and $\hat{\mathbf{x}}_0^2$ are predicted $\mathbf{x}_0$ of a single step for $\mathbf{x}_t^1$ and $\mathbf{x}_t^2$, respectively~\citep{song2020denoising}.

\textbf{Pairwise comparison}. After pair construction, at step $t$, we utilize a rank model denoted by $r(\cdot)$ to compare $\mathbf{x}_t^1$ and $\mathbf{x}_t^2$. This yields a preferred prediction $\mathbf{x}_t^{\text{win}}$ and a less preferred one $\mathbf{x}_t^{\text{lose}}$, where $\mathbf{x}_t^{\text{win}}=\mathbf{x}_t^1$ and $\mathbf{x}_t^{\text{lose}}=\mathbf{x}_t^2$, or vice versa. It is worth noting that our DreamDPO supports both reward models~\citep{xu2024imagereward,wu2023human} and LMM-based AI annotators~\citep{bai2023qwen,yang2023dawn}, where reward models are used as default.

\textbf{Preference-guided optimization.} 
The proposed method leverages the pairwise comparison
$(\mathbf{x}_t^{\text{win}},\mathbf{x}_t^{\text{lose}})$ to enable efficient sampling via optimization to yield human preferred 3D assets.
To achieve this, we need a differentiable loss function, where preferred images have low losses and less preferred images have high losses. To this end, inspired by~\citep{rafailov2024direct,meng2024simpo,wallace2024diffusion}, we reformulate SimPO~\citep{meng2024simpo} to eliminate the need for a reference model and derive a differentiable objective:
\begin{equation}
    \mathcal{L}_{\text{Reward}} = -\mathbb{E}_{t} \left[w(t)\left(  \| \bm{\epsilon}^\text{win} - \bm{\epsilon}_\phi(\mathbf{x}_t^\text{win};y,t) \|_2^2 
    - \| \bm{\epsilon}^\text{lose} - \bm{\epsilon}_\phi(\mathbf{x}_t^\text{lose};y,t) \|_2^2 \right) \right],
\end{equation}
where $\bm{\epsilon}^\text{win}$ and $\bm{\epsilon}^\text{lose}$ denote Gaussian noise for $\mathbf{x}_t^\text{win}$ and $\mathbf{x}_t^\text{lose}$ respectively. 
Intuitively, $\mathcal{L}_{\text{Reward}}$ encourages $\bm{\epsilon}_\phi$ to pull $\mathbf{x}_t^\text{win}$ closer and push $\mathbf{x}_t^\text{lose}$ further away.

Following~\citep{poole2022dreamfusion}, we consider the gradient of $\mathcal{L}_{\text{Reward}}$ and omit the U-Net Jacobian term for effective optimization.
It leads to the following gradient for optimizing 3D representations with preference pairwise comparisons:
\begin{equation}\label{eq:gradient_reward_loss}
    \nabla_\theta \mathcal{L}_{\text{Reward}}  = 
    \mathbb{E}_{t} \left[ w(t) \left( \bm{\epsilon}_\phi(\mathbf{x}_t^{\text{win}}; y, t) - \bm{\epsilon}^{\text{win}} \right)
    - \left( \bm{\epsilon}_\phi(\mathbf{x}_t^{\text{lose}}; y, t) - \bm{\epsilon}^{\text{lose}} \right) \frac{\partial \mathbf{x}}{\partial \theta} \right].   
\end{equation}
However, in practice, the gradient in \cref{eq:gradient_reward_loss} fails to produce realistic results (refer to \cref{fig:abl_score_gap}).
Delving into the optimization process, we observe that the pairwise comparison results can be overly similar, leading to nearly equal scores.
In this case, directly pushing $\mathbf{x}_t^\text{lose}$ away leads to chaotic gradients.
To address this, we introduce a piecewise optimization loss that selectively pulls $\mathbf{x}_t^\text{win}$ when the preference score gap $s_\text{gap}$ is small.
The gradient of the final loss is defined as:
\begin{equation}
    \nabla_\theta \mathcal{L}_{\text{Reward*}} :=
\begin{cases}
\mathbb{E}_{t} \left[ w(t) \left( \bm{\epsilon}^s_\phi(\mathbf{x}_t^{\text{win}}; y, t) \right) \frac{\partial \mathbf{x}}{\partial \theta} \right], & s_\text{gap} < \tau, \\
\mathbb{E}_{t} \left[ w(t) \left( \mathbf{\Delta}_t^{\text{win}} - \mathbf{\Delta}_t^{\text{lose}}
\right) \frac{\partial \mathbf{x}}{\partial \theta} \right], & \text{otherwise}.
\end{cases}
\label{eq:reward}
\end{equation}
where $\tau=0.001$ is a pre-defined threshold, $\mathbf{\Delta}_t^{\text{win}} :=\bm{\epsilon}^s_\phi(\mathbf{x}_t^{\text{win}}; y, t) - \bm{\epsilon}^{\text{win}}$, and $\mathbf{\Delta}_t^{\text{lose}} := \bm{\epsilon}^1_\phi(\mathbf{x}_t^{\text{lose}}; y, t) - \bm{\epsilon}^{\text{lose}}$.
Note that $s_\text{gap} := r(\mathbf{x}_t^{\text{win}}, y) - r(\mathbf{x}_t^{\text{lose}}, y)$ indicates the discrepancy of preference scores between $\mathbf{x}_t^{\text{win}}$ and $\mathbf{x}_t^{\text{lose}}$. 
Instruction questions can also be incorporated into LMM-based ranking models to provide explicit guidance (see empirical evaluations in \S\ref{sec:refined_analysis}). 



\begin{figure*}[t]
    \centering
    \includegraphics[width=1.0\linewidth]{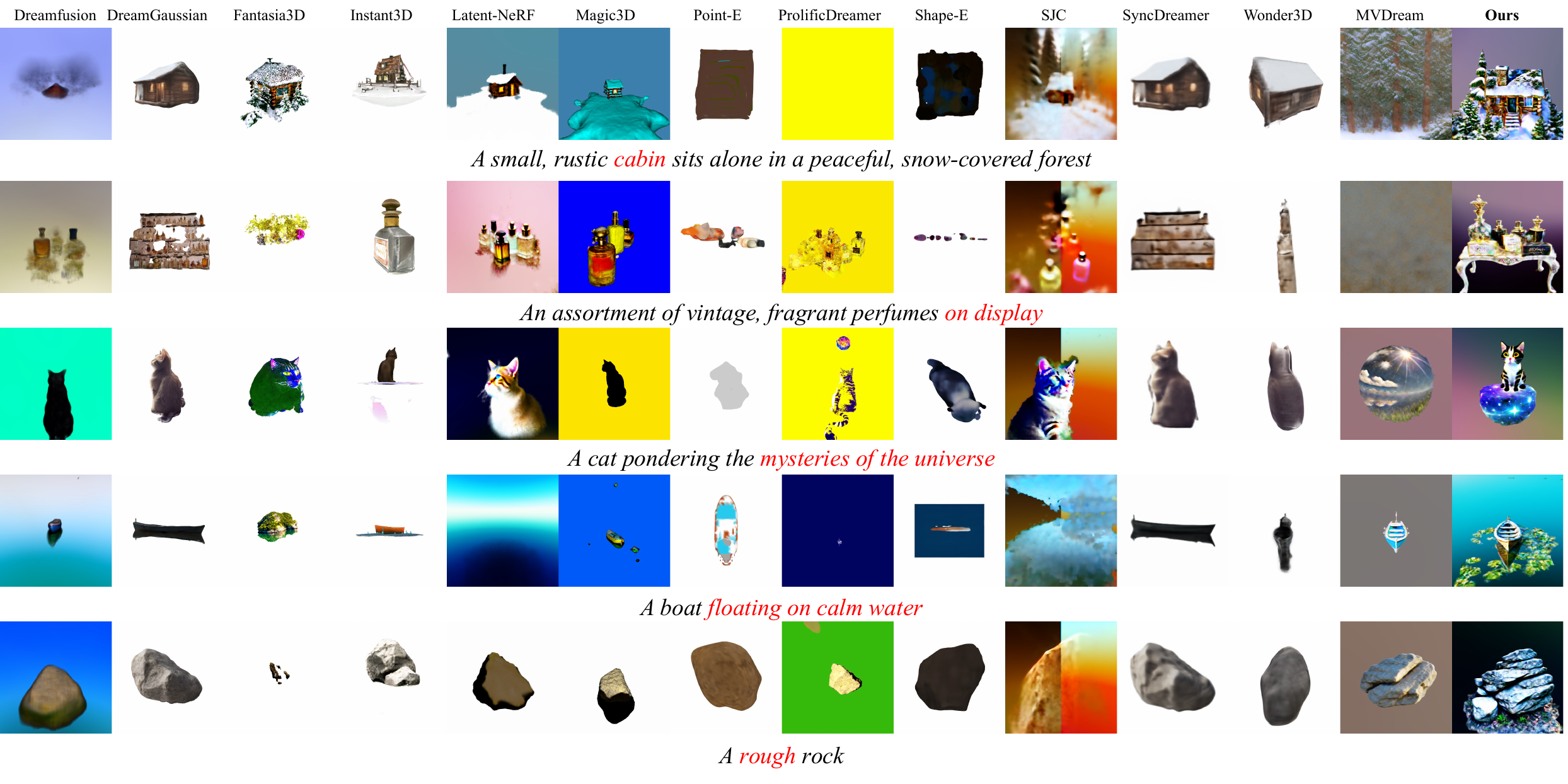}
    \caption{
    Qualitative comparisons on the benchmark of GPTEval3D~\protect\citep{wu2024gpt}.
    Existing methods struggle with text matching, as marked in red.
    DreamDPO improves text matching, which provides better human preference results.
    (Zoom in to see the details.)
    }
    \label{fig:quan_comp_all}
\end{figure*}

\section{Experiments}
\label{sec:exp}
In this section, a series of experiments are conducted to justify our claims. We first detail experiment setups~(\S\ref{sec:exp_setups}). The comprehensive results and comparison with previous advanced methods are then presented and discussed~(\S\ref{sec:exp_comparison}). Finally, we carry out an analysis study to further elaborate on and discuss the superiority of our method~(\S\ref{sec:refined_analysis}).

\begin{figure}[t]
    \centering
    \includegraphics[width=1.0\linewidth]{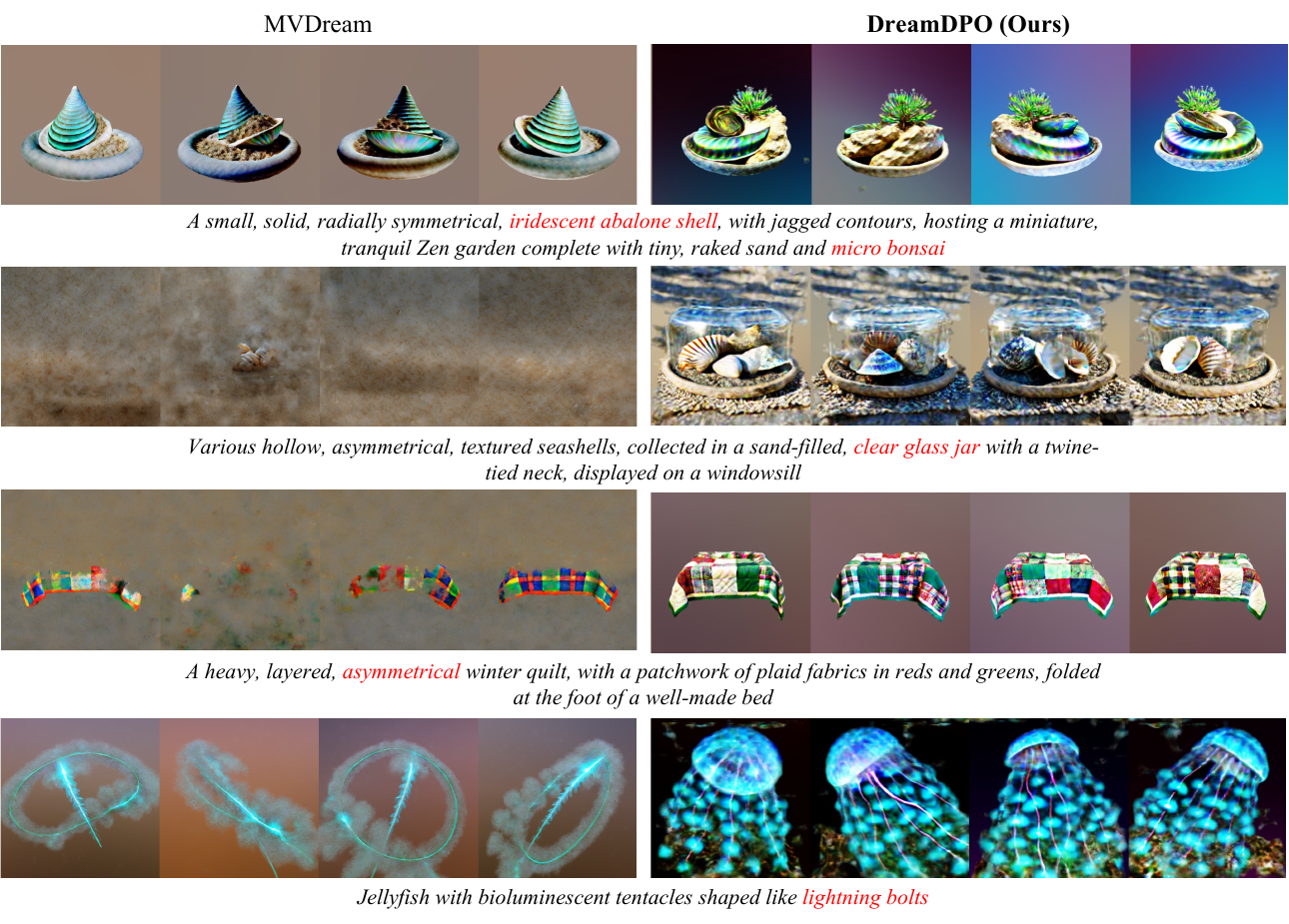}
    \caption{
    Qualitative comparisons with MVDream~\protect\citep{shi2023mvdream}. 
    DreamDPO performs well across short to long prompts, offering better human preference results, marked in red.~(Zoom in to see the details.)
    }
    \label{fig:quan_comp_backbone}
\end{figure}

\subsection{Experimental Setups}\label{sec:exp_setups}

\textbf{Datasets and measurements.} We here evaluate the proposed method with 110 prompts from GPTEval3D~\citep{wu2024gpt}, which covers a range of creativity and complexity use cases. Based on this, two evaluation strategies are exploited. 
(1) We utilize a text-to-image reward model named ImageReward~\citep{xu2024imagereward} to evaluate human preference for 3D assets.
We calculate the average preference score across 120 rendered images of a 3D asset and its corresponding text prompt.
(2) We use GPT-4V to perform pairwise comparisons with baselines, generating Elo ratings that align with human judgments on text alignment, 3D plausibility, and texture-geometry coherence, \etc. More details of the two measurements can be found in Appendix~\ref{app:metrics}.

\textbf{Baselines.} Following GPTEval3D~\citep{wu2024gpt}, we benchmark our method against 13 baselines categorized into text-guided and image-guided approaches respectively. Specifically, the text-guided group includes DreamFusion~\citep{poole2022dreamfusion}, DreamGaussian~\citep{tang2023dreamgaussian}, Instant3D~\citep{li2023instant3d}, Fantasia3D~\citep{chen2023fantasia3d}, Latent-NeRF~\citep{metzer2023latent}, Magic3D~\citep{lin2023magic3d}, MVDream~\citep{shi2023mvdream}, Point-E~\citep{nichol2022point}, ProlificDreamer~\citep{wang2024prolificdreamer}, Shap-E~\citep{jun2023shap}, and SJC~\citep{wang2023score}. Besides, the image-guided group includes SyncDreamer~\citep{liu2023syncdreamer} and Wonder3D~\citep{long2024wonder3d}.

\textbf{Implementation.} We conduct experiments using PyTorch~\citep{paszke2019pytorch} and threestudio~\citep{threestudio2023}, with MVDream~\citep{shi2023mvdream} as the backbone of our method. Note that we use PyTorch auto-differentiation to compute analytic normals for geometry evaluation in GPTEval3D and do not use the Lambertian shading trick~\citep{lin2023magic3d} due to memory limitation. We follow the training strategy of MVDream and use HPSv2~\citep{wu2023human} as the default reward model.
The optimization process takes around two hours on a single NVIDIA RTX A6000 GPU.


\subsection{Comparison with Prior Methods}\label{sec:exp_comparison}

\subsubsection{Qualitative Comparisons}
\begin{figure*}[t]
    \centering
    \includegraphics[width=1.0\linewidth]{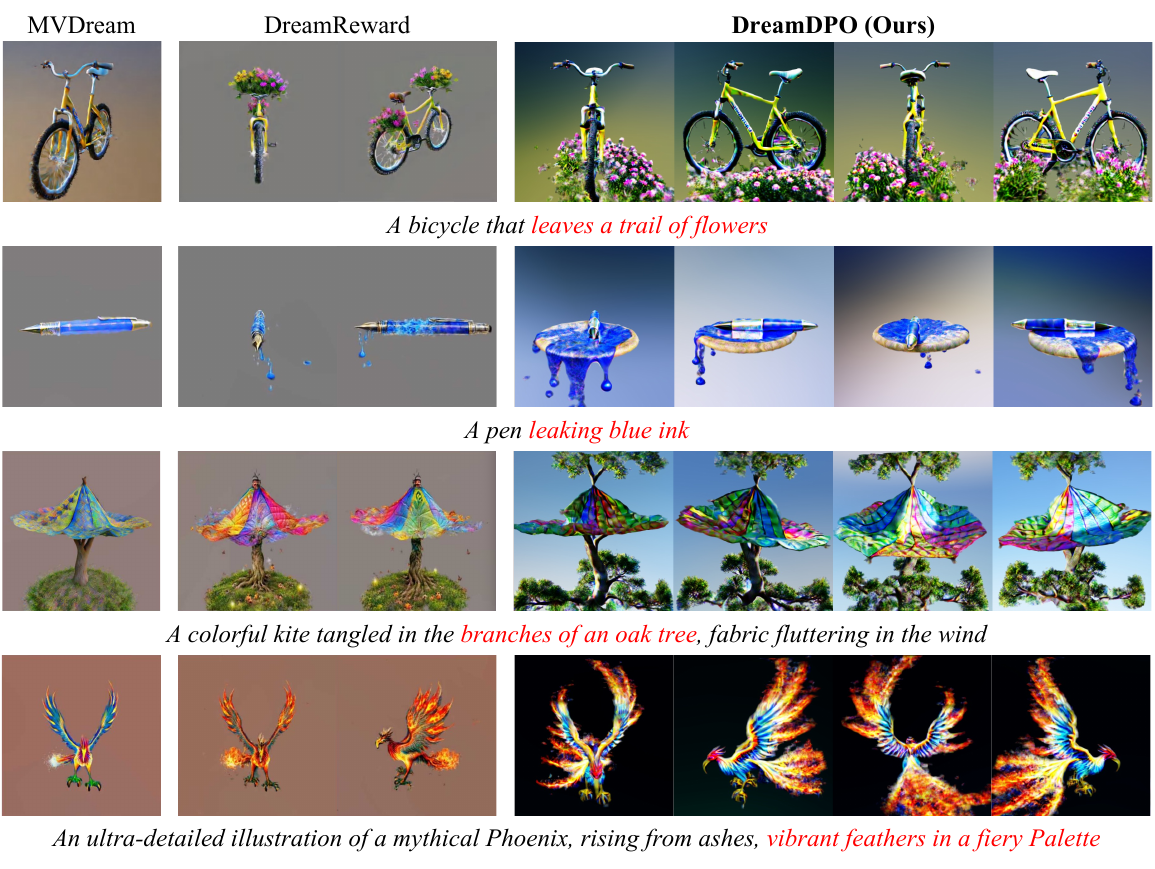}
    \caption{
    Qualitative comparisons with DreamReward~\citep{ye2025dreamreward}.
    DreamDPO improves both text matching (marked in red) and geometric/texture details.
    }
    \label{fig:quan_comp_baseline}
\end{figure*}

We conduct two qualitative evaluations, which include comparing 13 benchmarks of GPTEval3D, MVDream~\citep{shi2023mvdream}, and DreamReward~\citep{ye2025dreamreward}. The evaluations show improvements in text alignment, generation stability, and texture-geometry details, respectively.
As seen in \cref{fig:quan_comp_all}, while the comparing baselines produce high-fidelity results, they often fail in text alignment, as marked in red.
For instance, in the prompt ``\textit{A small, rustic cabin sits alone in a peaceful, snow-covered forest}'', most existing methods miss key elements like the forest (the first row in \cref{fig:quan_comp_all}).
In contrast, our method accurately captures both objects, showcasing its effectiveness for improving text alignment.
Further comparisons with MVDream, are shown in \cref{fig:quan_comp_backbone}.
Although MVDream is capable of generating multiview consistent 3D assets, it struggles with long prompts (\eg, the second and fourth rows in \cref{fig:quan_comp_backbone}).
Instead, our method performs well across both short to long prompts.
Lastly, the comparisons with DreamReward demonstrate that our method not only improves text alignment (\eg, ensuring ``\textit{leaves a trail of flowers}'' appears under the bicycle shown in the first row in \cref{fig:quan_comp_baseline}) but also enhances geometric and texture details (\eg, generating a more luxuriant oak shown in the second row in \cref{fig:quan_comp_baseline}).
More visualization results can be found in Appendix~\ref{app:qualitative_results}.

\subsubsection{Quantitative Comparisons}
We provide extensive quantitative comparison results to justify our claims.
Human preference evaluations are first conducted using ImageReward, as illustrated in the first column of \cref{tab:qua_comp}.
Our method achieves competitive performance compared to existing methods, highlighting the benefits of preference-guided optimization via our reward loss.
Then we perform a comprehensive evaluation using GPTEval3D. The results indicate that our method outperforms previous state-of-the-art (SOTA) methods, and ranks first across all metrics. Specifically, our method achieves improvements in text-asset alignment (+28.4), 3D plausibility ($+24.4$), text-geometry alignment (+25.8), texture details (+48.4), geometry details (+41.4), and overall performance (+24.4). 
It showcases the superiority of the proposed method in enhancing text and geometry details while maintaining 3D consistency.

\begin{figure}[t]
  \centering
  \begin{minipage}{0.5\textwidth}
    \centering
    \includegraphics[width=1.0\textwidth]{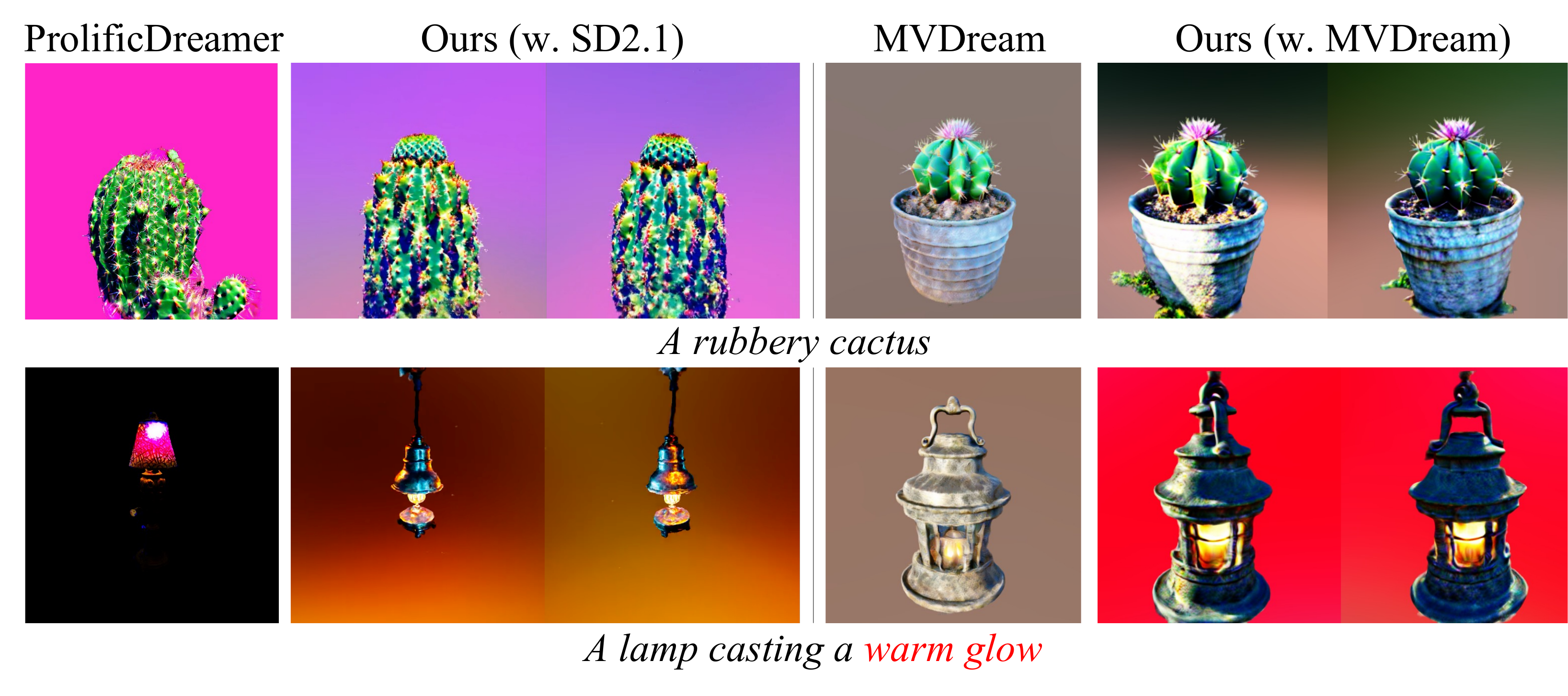}
    \caption{The analysis of backbone.
    We present the results of DreamDPO using Stable Diffusion v2.1 (SD2.1)~\citep{rombach2022high}.
    DreamDPO demonstrates effective performance with SD2.1, highlighting its potential to leverage more advanced backbone diffusion models for further improvements.}
    \label{fig:abl_backbone}
  \end{minipage}\hfill
  \begin{minipage}{0.48\textwidth}
    \centering
    \includegraphics[width=1.0\textwidth]{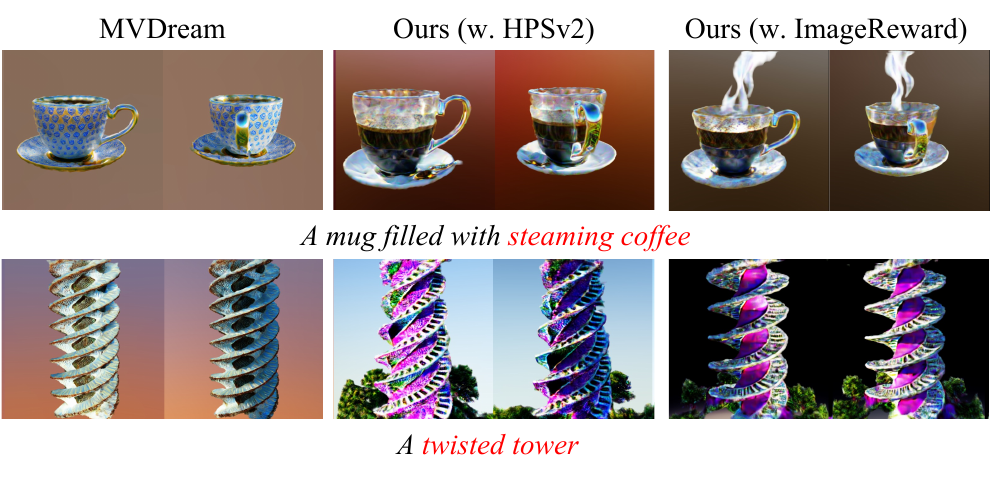}
    \caption{The analysis of reward models.
    We present the results of DreamDPO using ImageReward~\citep{xu2024imagereward}.
    DreamDPO demonstrates effective performance with ImageReward, highlighting its potential to leverage stronger reward models to further enhance generation quality.}
    \label{fig:abl_imagereward}
  \end{minipage}
\end{figure}

\begin{figure}[t]
    \centering    
    \includegraphics[width=1.0\linewidth]{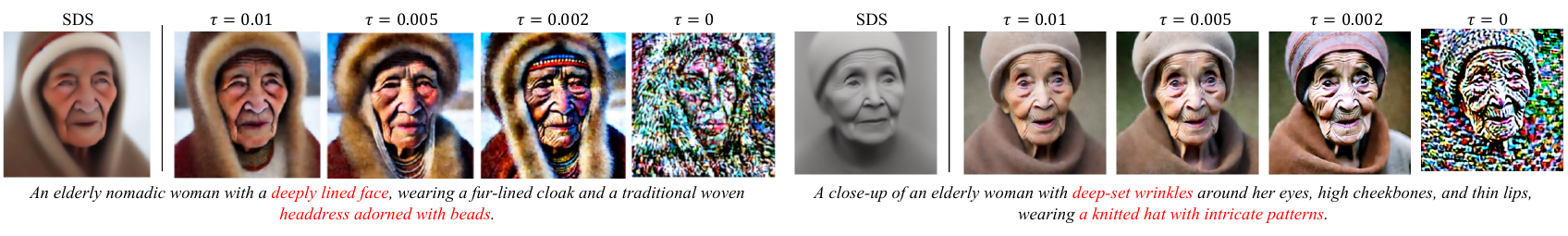}     
    \caption{
    The analysis of the score gap threshold $\tau$.     
    We conduct 2D toy experiments with $\tau$ ranging from $0.01$ to $0$.
    The results indicate that a small but non-zero $\tau$ effectively filters out overly similar \texttt{lose} examples, leading to more detailed outputs.
    }   
    \label{fig:abl_score_gap} 
\end{figure}

\begin{figure*}[t]
    \centering
    \includegraphics[width=1.0\linewidth]{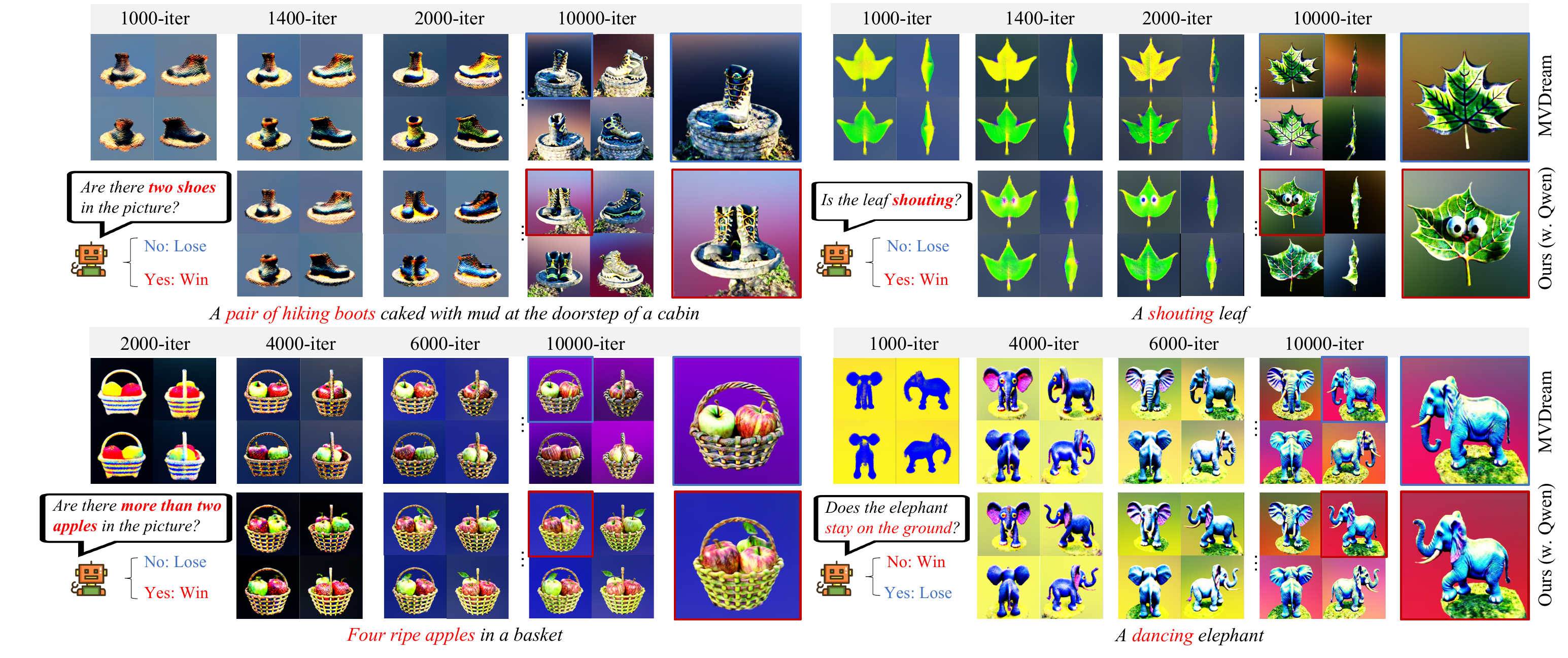}
    \caption{
    The generation results of DreamDPO with large multi-modal models (LMMs).
    We explore the potential of our method to leverage LMMs, such as QwenVL~\protect\citep{bai2023qwen} for explicit guidance in correcting the number and attribute of 3D assets. The left corner shows the details of pairwise comparisons using the LMM, including the question and win/lose criteria.
    By carefully designing the question, DreamDPO can leverage both \texttt{win} and \texttt{lose} examples to guide optimization. (Zoom in to see the details.)
    }
    \label{fig:abl_qwenv2}
\end{figure*}

\begin{figure}[t]
  \centering
  \begin{minipage}{0.5\textwidth}
    \centering
    \includegraphics[width=1.0\textwidth]{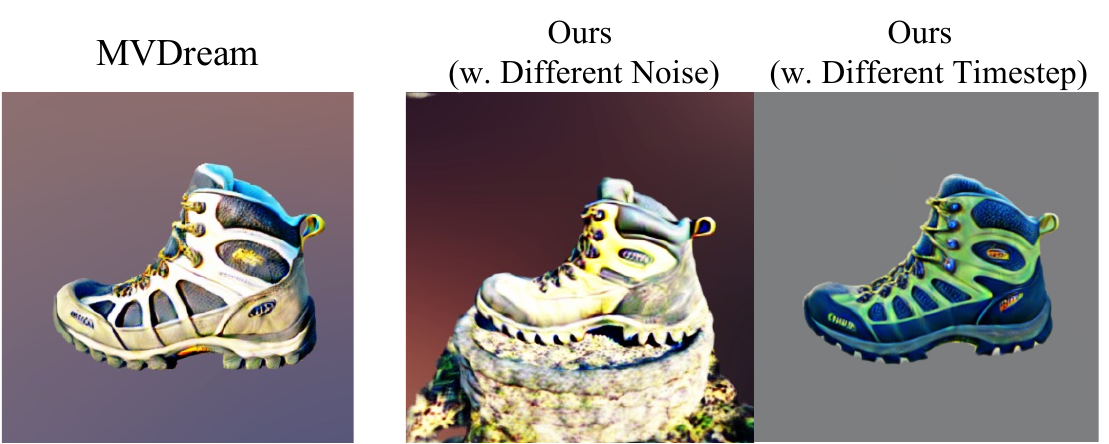}
    \caption{The analysis of pairwise example construction.
    We compare (1) different noises: adding different Gaussian noises with the same timesteps, and (2) difference timesteps: adding the same Gaussian noise with different timesteps.}
    \label{fig:abl_negative_example}
  \end{minipage}\hfill
  \begin{minipage}{0.48\textwidth}
    \centering
    \includegraphics[width=1.0\textwidth]{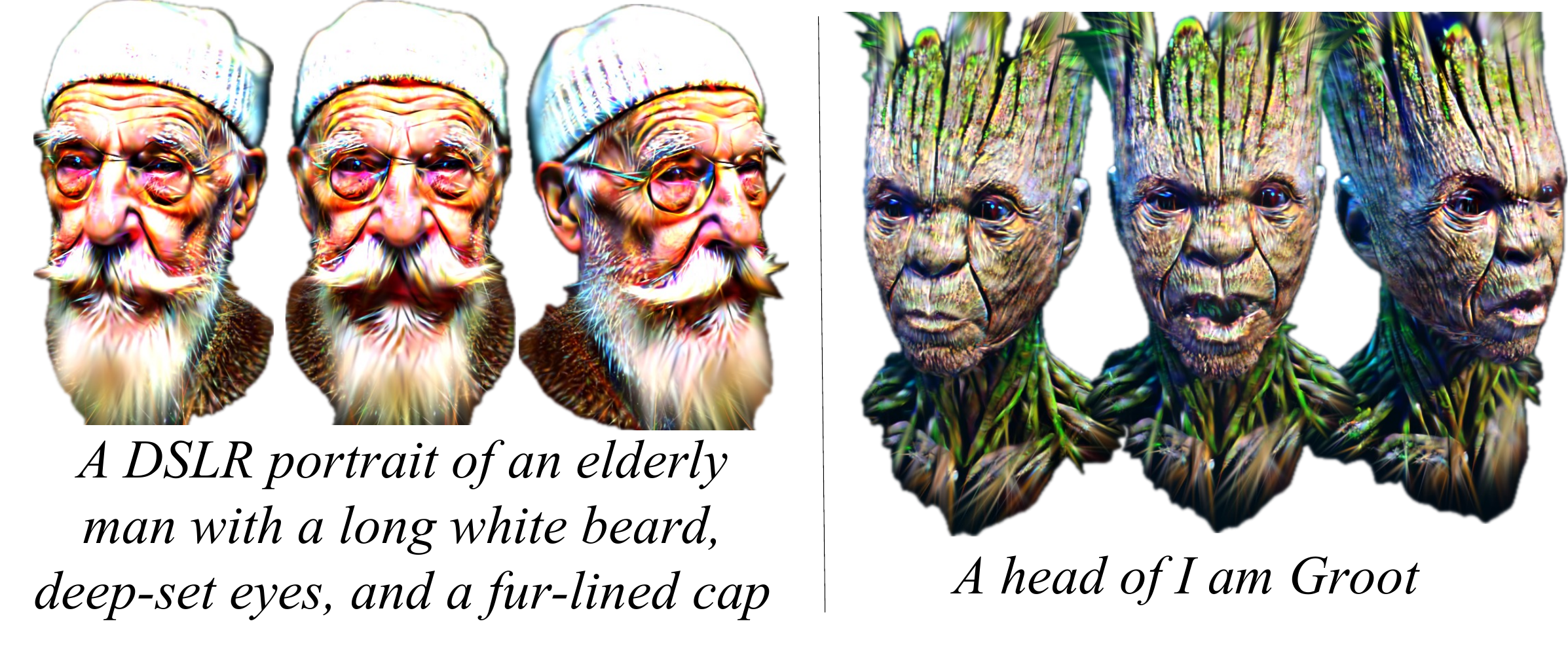}
    \caption{The further application of DreamDPO.
    We conduct toy experiments on text-to-avatar generation by combining DreamDPO with Gaussian-based avatar generation framework~\citep{zhou2024headstudio}. More details can be checked in Appendix~\ref{app:avatar}.}
    \label{fig:application}
  \end{minipage}
\end{figure}

\subsection{More Analyses and Justifications}\label{sec:refined_analysis}

\textbf{Evaluation on different backbones. }
We here further investigate the impact of backbone selection on our method. The performance of DreamDPO using Stable Diffusion v2.1~(SD2.1)~\citep{rombach2022high} is provided. Note that as the previous method ProlificDreamer~\citep{wang2023prolificdreamer} also utilizes SD2.1, we compare DreamDPO with SD2.1 against ProlificDreamer. As shown in~\cref{fig:abl_backbone}, our method can perform effectively with SD2.1 and achieve competitive results compared to ProlificDreamer. Importantly, DreamDPO does not need LoRA training and is \textit{more efficient} than ProlificDreamer. 

\textbf{Evaluation on different reward models.} We study the impact of reward model selection on our method. Specifically, ImageReward~\citep{xu2024imagereward} is used, which is an image-based reward model with a similarity function comparable to HPSv2~\citep{wu2023human}. As shown in \cref{fig:abl_imagereward}, the results demonstrate that our method performs effectively across different reward models, demonstrating its flexibility and scalability. For instance, in the prompt ``\textit{A mug filled with steaming coffee}'', both HPSv2 and ImageReward successfully capture the coffee, and ImageReward places greater emphasis on the steam.
While ImageReward demonstrates improvement over the baseline, HPSv2 yields superior results due to its better generalization across diverse image distributions~\citep{wu2023human}. 
Therefore, we adopt HPSv2 as our default reward model. Also, it highlights the potential of leveraging stronger reward models, \textit{e.g.}, Reward3D~\citep{ye2025dreamreward} and VisionReward~\citep{xu2024visionreward}, to further enhance generation quality.

\textbf{Evaluation on different score gaps.} We investigate the impact of the score gap $\tau$. Specifically, 2D toy experiments with $\tau$ range from $0.01$ to $0$ are conducted. We provide results in~\cref{fig:abl_score_gap}, which show that a smaller $\tau$ produces more detailed outputs. Note that a small $\tau$ means choosing to \texttt{lose} examples with scores close to \texttt{win} samples, and focusing the training process on hard cases. However, we observe that $\tau = 0$ (the last column in \cref{fig:abl_backbone}) results in a chaotic gradient.
To balance high-fidelity generation and stable training, we suggest using \textit{a small but non-zero} $\tau$, which excludes overly similar \texttt{lose} examples.

\textbf{Evaluation on different pair examples.} The influence of different pair example generation methods is studied. Specifically, we compare: (1) \textit{different noises}, by adding different Gaussian noises with the same timesteps; (2) \textit{difference steps}, by adding the same Gaussian noise with different timesteps. As shown in~\cref{fig:abl_negative_example}, using different Gaussian noise yields better results than different timesteps. We attribute that noisy latents with different timesteps are easier to distinguish, making them less effective as challenging examples.
It highlights the importance of generating meaningful and challenging \texttt{lose} examples.  Accordingly, we adopt different noises as the default setting.

\textbf{Ranking model design.} We explore the potential of our method to leverage large multi-modal models (LMMs) for explicit guidance. Instead of relying on a reward model, we use large visual-language models, such as QwenVL~\citep{bai2023qwen}, to rank paired results. Specifically, we extract ``yes'' or ``no'' questions from the text prompt, query the LMM with rendered paired examples, and count the ``yes'' responses to calculate the reward score. 
Then, we use \cref{eq:reward} with threshold $\tau = 1$ for 3D assets generation.
As shown in \cref{fig:abl_qwenv2}, our method effectively improves text alignment by using LMM to guide the optimization with user instruction (\eg, correcting the \textit{number} and \textit{attribute} of 3D assets). Additionally, our method flexibly supports using \texttt{lose} examples to guide optimization. 
For example, given the prompt ``\textit{A dancing elephant}'', the baseline generates an elephant standing rather than dancing. 
By setting ``\textit{elephant stays on the ground}'' as the \texttt{lose} example and pushing it away, our method encourages the elephant to lift its leg, leading to a dancing pose. It highlights the potential of our method to integrate LMMs into 3D generation.
More implementation details of the LMMs-based comparison can be found in Appendix~\ref{app:lmms}.

\textbf{Further application.} 
To showcase the potential of our method, we provide empirical results on text-to-avatar generation. In specific, we replace the score sampling loss in HeadStudio~\citep{zhou2024headstudio} which is a 3DGS~\citep{kerbl3Dgaussians} based avatar generation framework, with our reward loss. 
As illustrated in \cref{fig:application}, our method achieves great generation results. This underscores the broader applicability of our method to various generation tasks, \eg, 4D generation~\citep{bahmani20244d} and scene generation~\citep{zhang2024text2nerf}, \etc.

\section{Conclusion}
\label{sec:clu}

In this work, we propose DreamDPO, an optimization-based 3D generation method that offers human preferences and fine-grained control for generation.
The method is built on three key steps: pairwise example construction, pairwise example comparison, and preference-guided optimization.
Unlike existing methods that rely on precise pointwise quality evaluations, DreamDPO uses pairwise comparison informed by reward models or large multimodal models. It enables a more flexible optimization process.
By incorporating human preferences directly into the optimization, DreamDPO generates 3D assets that are better aligned with input text and exhibit enhanced texture and geometry quality.
Comprehensive experimental results demonstrate that DreamDPO surpasses previous state-of-the-art methods in both output quality and controllability. Lastly, we hope DreamDPO paves the way for more refined, adaptable, and human-aligned 3D content generation solutions.

\textbf{Limitations and future work.}
While DreamDPO has shown improvements in aligning 3D generation with human preferences, several avenues for future research could further enhance its performance and applicability.
The primary limitations of DreamDPO are as follows:
(1) AI feedback can be used for guidelines but is largely limited to the inherent power of generative models. 
(2) Open API can provide more freedom but actually bring more instability, where instruction prompts should be designed carefully. 

To address these limitations, we suggest the following directions for future work:
(1) Enhancing generative models.
Incorporating image prompts~\citep{chen2024vp3d} to introduce explicit guidance could improve alignment with user expectations by providing a more detailed context for generation. (2) Improving the robustness of models in pairwise comparison.
Exploring prompt-free methods, such as leveraging object detection models~\citep{wang2023detecting} or grounding models~\citep{oquab2023dinov2}, for number and attribute correction, might reduce dependencies on prompt design. 
Additionally, using the diffusion model itself as a model for pairwise comparison~\citep{tian2024diffuse} could enhance stability and performance by ensuring consistency across the generation and comparison.

\bibliography{main}
\bibliographystyle{unsrt}

\appendix
\clearpage
\onecolumn
\appendix

\etocdepthtag.toc{mtappendix}
\etocsettagdepth{mtchapter}{none}
\etocsettagdepth{mtappendix}{subsection}

\renewcommand{\contentsname}{Appendix}
\tableofcontents

\clearpage

\section{Related Work}\label{app:related_work}
\subsection{Text-to-Image Generation}
With the development of vision-language models~\citep{radford2021clip} and diffusion models~\citep{sohl2015deep, ho2020denoising}, great advancements recently have been made in text-to-image generation~\citep{nichol2021glide, ho2022imagen, zhang20232dsurvey}. In particular, Stable Diffusion~\citep{stable_diffusion} is a notable framework that trains the diffusion models on latent space, leading to reduced complexity and detailed preservation. In addition, with the emergence of text-to-2D models, more applications have been developed, \eg, spatial control~\citep{voynov2023sketch, zhang2023controlnet}, concept control~\citep{gal2022image, ruiz2022dreambooth}, and image editing~\citep{brooks2023instructpix2pix}.

\subsection{Text-to-3D Generation}
The success of the 2D generation is incredible.
However, it is challenging to transfer image diffusion models to 3D because of the difficulty of 3D data collection. 
Fortunately, Neural Radiance Fields (NeRF)~\citep{mildenhall2020nerf, barron2022mip} provided new insights for 3D-aware generation, where only 2D multi-view images are needed in 3D scene reconstruction. 
Combining prior knowledge from text-to-2D models, several methods, such as DreamField~\citep{jain2021dreamfields}, DreamFusion~\citep{poole2022dreamfusion}, and SJC~\citep{wang2022sjc}, have been proposed to generate 3D objects guided by text prompts~\citep{li20233dsurvey}.
However, the vanilla score distillation sampling loss~\citep{poole2022dreamfusion} suffers from issues such as over-saturation, over-smoothing, and Janus problems, \etc. 
Recently, several works have proposed improvements to enhance generation quality~\citep{wang2023prolificdreamer, yu2023text, zhu2023hifa, katzir2023noise, chung2023luciddreamer, wu2024consistent3d, zhuo2025vividdreamer}.
Additionally, with the availability of large 3D datasets~\citep{deitke2023objaverse, deitke2024objaverse}, some works~\citep{liu2023zero, shi2023mvdream, liu2024one, liu2023syncdreamer, long2024wonder3d} leverage multi-view information to address the Janus problem more effectively.
Moreover, the recent advancement of text-to-3D generation also inspired multiple applications that include but are not limited to text-guided scene generation~\citep{cohen2023set, hollein2023text2room}, text-guided 3D editing~\citep{Ayaan2023instructnerf, kamata2023instruct3d}, and text-guided avatar generation~\citep{cao2023dreamavatar, jiang2023avatarcraft,han2023headsculpt,zhou2024headstudio}.

\subsection{Learning from Human Preferences}
Learning from human preferences is essential for improving the alignment and performance of generative models across various domains, including large language models (LLMs)~\citep{bai2022training,bai2022constitutional,leerlaif}, large multimodal models (LMMs)~\citep{2023llavarlhf,yu2024rlhf,yu2024rlaifv}, text-to-image diffusion models~\citep{black2023training,lee2023aligning,clark2023directly,xu2024imagereward,wallace2024diffusion,fan2024reinforcement,zhang2024hive}, and text-to-3D generation~\citep{xie2024carve3d,ye2025dreamreward}.
Reinforcement Learning from Human Feedback (RLHF)~\citep{christiano2017deep} has been proven effective in refining LLMs to better align with human preferences. 
It often involves collecting human feedback datasets, training a reward model, and fine-tuning the language model using reinforcement learning.
InstructGPT~\citep{ouyang2022training} is a notable work that employs a two-stage fine-tuning strategy to align GPT-3 with human instructions, which leads to more coherent and contextually appropriate outputs.

The success of RLHF in LLMs has inspired the applications in text-to-image diffusion models to enhance image generation quality and align with human preferences. In more detail, RWR~\citep{lee2023aligning} first introduces human feedback-based reward fine-tuning for diffusion models, which fine-tunes Stable Diffusion~\citep{rombach2022high} using log probabilities of the denoising process. 
ImageReward~\citep{xu2024imagereward} proposes a reward model specifically for text-to-image tasks and further develops reward feedback Learning for refining diffusion models. 
DiffusionDPO~\citep{wallace2024diffusion} uses DPO to optimize diffusion models using human comparative data, while 
DPOK~\citep{fan2024reinforcement} integrates policy optimization with KL regularization for improved alignment.
Recently, advancements in multi-view diffusion models have facilitated significant progress in text-to-3D generation. For instance,
Carve3D~\citep{xie2024carve3d} enhances text-to-3D generation with a Multi-view Reconstruction Consistency (MRC) metric for improved consistency and quality.
DreamReward~\citep{ye2025dreamreward} improves text-to-3D models using human feedback. 
It collects a 3D dataset with human annotations, trains Reward3D as a multi-view reward model, and introduces DreamFL to create 3D assets aligned with human preferences.
However, these works rely heavily on large-scale datasets to train a reward model or utilize it as preference feedback, which is very expensive for 3D generation.

\section{Additional Implementation Details}\label{app:algorithm_flow}

\subsection{Pseudo-Code for DreamDPO}
A more detailed pseudo-code for DreamDPO is presented in~\cref{alg:dreamdpo}.

\begin{algorithm}
    \renewcommand{\algorithmicrequire}{\textbf{Input:}}
    \renewcommand{\algorithmicensure}{\textbf{Output:}}
    \caption{Pseudo-code for DreamDPO}
    \label{alg:dreamdpo}
    \begin{algorithmic}[1]
        \REQUIRE Text-to-image diffusion model $\bm{\epsilon}_\phi$. Ranking model $r$. Learning rate $\eta$ for 3D representation parameters. A prompt $y$. Evaluating threshold of score gap $\tau$.
        \STATE Initialization A 3D representation presenting with NeRF $\theta$
        \WHILE{not converged}
        \STATE Randomly sample a camera pose $c$, pairwise 2D noise $\bm{\epsilon}^1$ and $\bm{\epsilon}^2$, and timestep $t \sim \text{Uniform}(\left\{ 1, \cdots, T\right\})$.
        \STATE Render at pose $c$ to get a image $\mathbf{x}_0$.
        \STATE Add noise $\bm{\epsilon}^1$ and $\bm{\epsilon}^2$ to $\mathbf{x}_0$ and get $\mathbf{x}_t^1$ and $\mathbf{x}_t^2$, respectively.
        \STATE Denoise with the predicting noise:
        \begin{equation} 
    \begin{aligned}
       \hat{\mathbf{x}}_0^1 = \frac{\mathbf{x}_t^1 - \sqrt{1 - \alpha_t}\bm{\epsilon}_\theta(\mathbf{x}_t^1;y,t)}{\sqrt{\alpha_t}}, \\
       \hat{\mathbf{x}}_0^2 = \frac{\mathbf{x}_t^2 - \sqrt{1 - \alpha_t}\bm{\epsilon}_\theta(\mathbf{x}_t^2;y,t)}{\sqrt{\alpha_t}}.
       \nonumber
    \end{aligned}
\end{equation}
        \STATE Score the prediction $(\hat{\mathbf{x}}_0^1, \hat{\mathbf{x}}_0^2)$ online via a rank model $r$, yielding the pairwise comparison $(\mathbf{x}_t^{\text{win}}, \mathbf{x}_t^{\text{lose}})$.
        \STATE Compute the score gap:
        \begin{equation}\notag
            s_\text{gap} = r(\mathbf{x}_t^{\text{win}}, y) - r(\mathbf{x}_t^{\text{lose}}, y).
        \end{equation}
        \IF{$s_\text{gap} < \tau$ }
        \STATE
        \begin{equation}
            \nabla_\theta \mathcal{L}_{\mathrm{Reward}} = \mathbb{E}_{t} \left[ w(t) \left( \bm{\epsilon}^s_\phi(\mathbf{x}_t^{\text{win}}; y, t) \right) \frac{\partial \mathbf{x}}{\partial \theta} \right],
            \nonumber
        \end{equation}
        \ELSE
        \STATE 
        \begin{equation}
            \nabla_\theta \mathcal{L}_{\mathrm{Reward}} = \mathbb{E}_{t} \left[ w(t) \left( \left( \bm{\epsilon}^s_\phi(\mathbf{x}_t^{\text{win}}; y, t) - \bm{\epsilon}^{\text{win}} \right) - \left( \bm{\epsilon}^1_\phi(\mathbf{x}_t^{\text{lose}}; y, t) - \bm{\epsilon}^{\text{lose}} \right) \right) \frac{\partial \mathbf{x}}{\partial \theta} \right].
            \nonumber
        \end{equation}
        \ENDIF
        \item $\theta \leftarrow \theta - \eta \nabla_\theta \mathcal{L}_{\mathrm{Reward}}$.
        \ENDWHILE
    \end{algorithmic}
\end{algorithm}

\subsection{Details of LMM-based Pairwise Comparison}
\label{app:lmms}
We detail the implementation of the LMM-based pairwise comparison.
We use the large visual-language model ``qwen-vl-plus-latest'' from QwenVL~\citep{bai2023qwen} as the default LMM.
Given pairwise examples, we conduct the comparison query sequentially.
For each query, we first insert a predefined ``yes" or ``no" question into the comparison prompt, such as ``\textit{Is the leaf shouting?}'' for the prompt ``\textit{A shouting leaf.}'' 
Then, the LMM performs visual question answering based on the provided image and query. 
Finally, we extract the number of ``yes" responses as the score.
The following prompts are used for the queries:
\begin{tcolorbox}[colback=blue!5!white, colframe=blue!75!black, 
    title=Comparison Query, width=\textwidth, 
    boxrule=1pt, arc=2mm, auto outer arc,
    sharp corners=south]
\textcolor{blue}{[Task Description]}: You are an expert in evaluating the alignment between a given text description and an image. Your task is to answer each of the alignment questions with either ``Yes" or ``No" based on the image. Provide your responses in the format specified below.

\textcolor{orange}{[Evaluation Instruction]:} 

1. Carefully analyze the provided image and answer questions based on the image.

2. For each question, answer with either ``Yes" or ``No". Do not provide explanations or additional information.

\textcolor{olive}{[Evaluation Question(s)]:} 

Q1: \{Question\}

...

\textcolor{magenta}{[Output Format]:}

A1: [Yes/No]

...
\end{tcolorbox}

\subsection{Details of Text-to-Avatar Generation}\label{app:avatar}
We detail the toy exploration of text-to-avatar generation using DreamDPO. 
Specifically, we integrate the reward loss into HeadStudio~\citep{zhou2024headstudio}, a Gaussian-based avatar generation framework. 
HeadStudio is an optimization-based method that utilizes a score sampling loss~\citep{katzir2023noise} with ControlNet~\citep{zhang2023controlnet} to optimize an animatable head prior model. 
By replacing the score sampling loss with the reward loss, we leverage ControlNet to generate avatars.

\section{Supplementary Experimental Settings}

\subsection{Details of Measurement Metrics}\label{app:metrics}
In the main paper, we employ two evaluation strategies to demonstrate the superiority of the proposed method. Here we supplementary the details of the measurements.

\textbf{Evaluation with ImageReward.}
ImageReward~\citep{xu2024imagereward} is a text-to-image human preference reward model.
Due to its effectiveness, it has been broadly used for human preference evaluation in text-to-image generation~\citep{fan2024reinforcement} and text-to-3D generation~\citep{ye2025dreamreward}.
Given a (text, image) pair, it extracts image and text features, combines them with cross-attention, and uses an MLP to generate a scalar for preference comparison.
For each 3D asset, we uniformly render 120 RGB images from different viewpoints.
Afterward, the ImageReward score is computed from the multi-view renderings and averaged for each prompt.

\textbf{Evaluation with GPTEval3D.}
We utilize GPTEval3D~\citep{wu2024gpt}, which is a comprehensive benchmark for text-to-3D generation evaluation. GPTEval3D includes 13 baseline methods $\mathcal{M}$, 110 text prompts, and 5 criteria that are text-asset alignment, 3D plausibility, texture details, geometry details, and texture-geometry coherency respectively.
For a new method, GPTEval3D employs GPT-4V to compare 3D assets generated by this new method and one of the baseline methods with the same input text prompt.
These pairwise comparison results are then used to calculate the Elo rating for each model.
Specifically, let $\mathbf{A}$ be a matrix where $\mathrm{A}_{ij}$ represents the number of times that the $i$-th model outperforms the $j$-th model in comparisons. 
The Elo ratings for the models are computed by optimizing the following objective:
\begin{equation}
    \sigma = \arg\max\limits_{\sigma} \mathop{\sum}\limits_{i \neq j} \mathrm{A}_{ij} \log \left(1 + 10^{(\sigma_j - \sigma_i)/400}\right),
\end{equation}
where $\sigma_i \in \mathbb{R}$ is the Elo rating of the $i$-th model.
In this work, we calculate Elo ratings within the existing tournament, initializing, and freezing baseline scores as specified in the official code\footnote{\href{https://github.com/3DTopia/GPTEval3D/blob/main/data/tournament-v0/config.json}{https://github.com/3DTopia/GPTEval3D/blob/main/data/tournament-v0/config.json}}. For interested readers, please refer to~\citep{wu2024gpt}.

\section{Supplementary Experimental Results}

\subsection{More Qualitative Results}\label{app:qualitative_results}
\begin{figure*}[ht]
    \centering
    \includegraphics[width=1.0\linewidth]{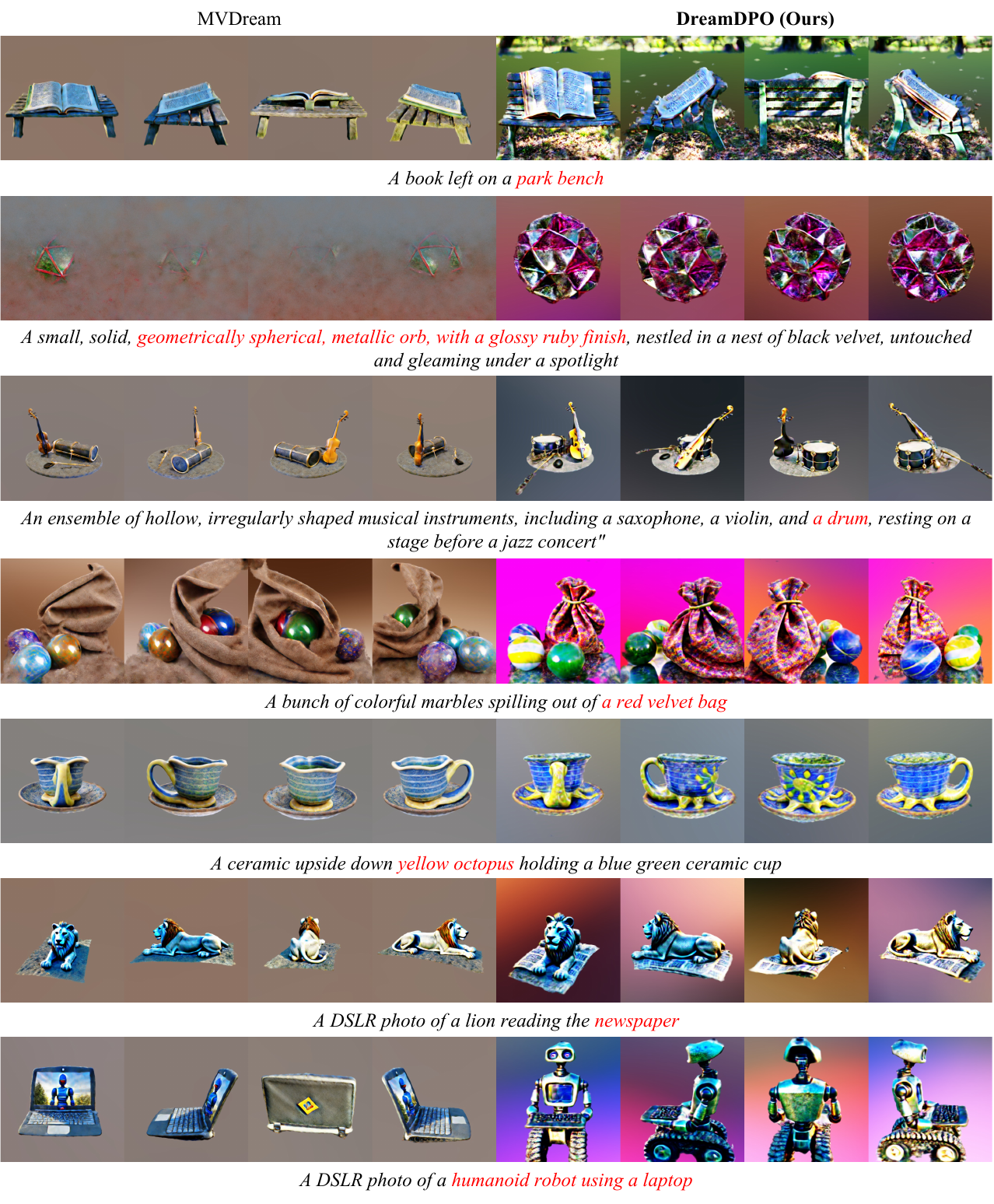}
    \caption{More qualitative results using DreamDPO.}
    \label{fig:supp_comp_mvdream}
\end{figure*}

\begin{figure*}[ht]
    \centering
    \includegraphics[width=1.0\linewidth]{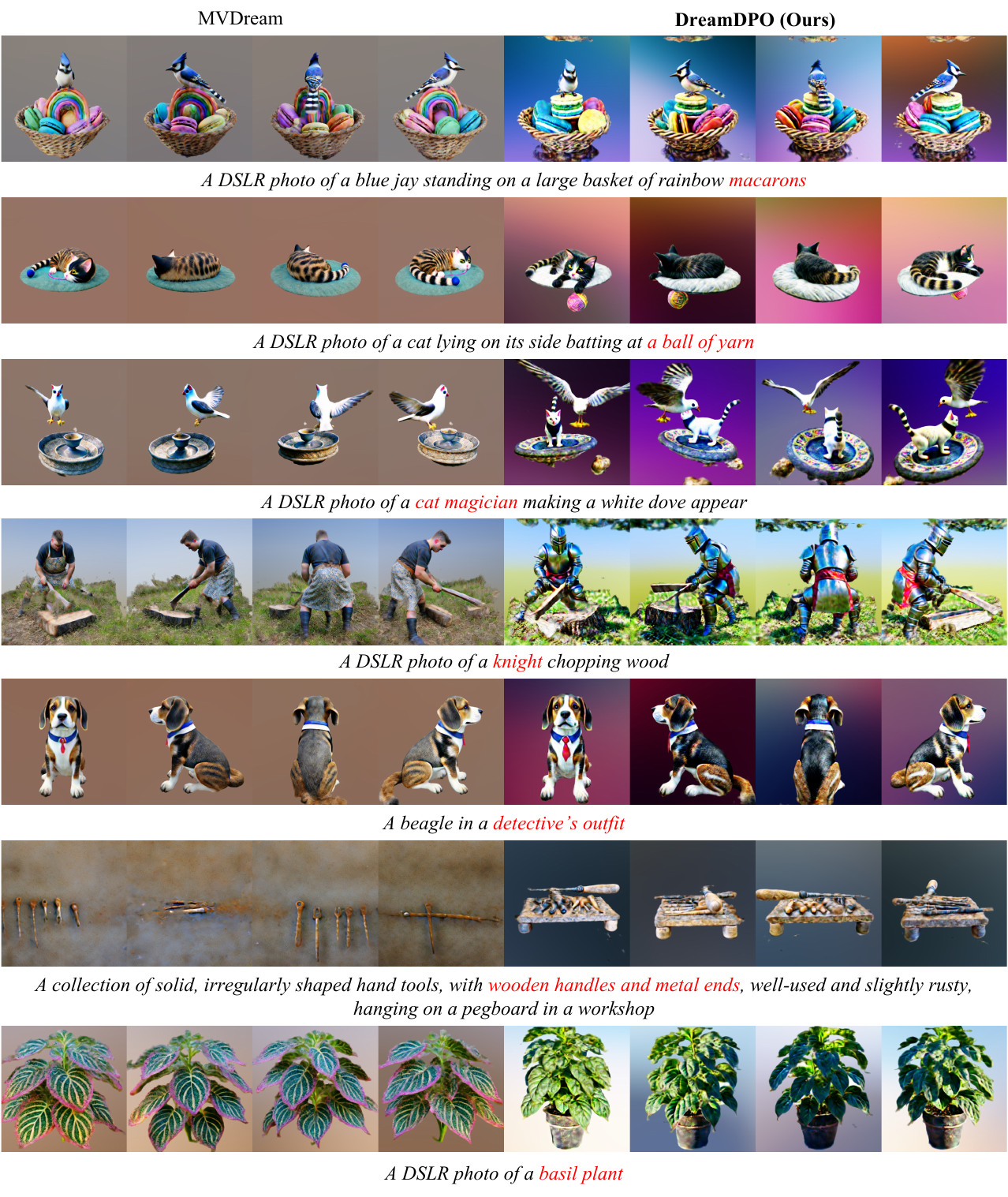}
    \caption{More qualitative results using DreamDPO.}
    \label{fig:supp_comp_mvdream2}
\end{figure*}

We present additional qualitative results in \cref{fig:supp_comp_mvdream} and \cref{fig:supp_comp_mvdream2}. 
The comparisons demonstrate that our method generates human preferred 3D assets, with improved text alignment and enhanced geometric and texture details.

\end{document}